\documentclass[sensors,article,accept,moreauthors,pdftex]{Definitions/mdpi} 

\usepackage{booktabs} 
\usepackage{multirow}
\usepackage{soul} 
\usepackage{microtype}

\firstpage{1} 
\pubvolume{21}
\issuenum{6}
\articlenumber{2032}
\pubyear{2021}
\copyrightyear{2021}
\externaleditor{Academic Editor: Enrico Meli} % For journal Automation, please change Academic Editor to "Communicated by"
\datereceived{19 February 2021} 
\dateaccepted{11 March 2021} 
\datepublished{13 March 2021} 
\hreflink{https://doi.org/10.3390/s21062032} % If needed use \linebreak
\usepackage{amssymb}
\usepackage{array}
\usepackage{url}
\usepackage{verbatim}
\usepackage[section]{placeins}
\Title{Weakly Supervised Reinforcement Learning for Autonomous Highway Driving via Virtual Safety Cages}

\TitleCitation{Weakly Supervised Reinforcement Learning for Autonomous Highway Driving via Virtual Safety Cages}

\Author{Sampo Kuutti 
 $^{1,}$*\orcidA{}, Richard Bowden $^{2}$\orcidB{} and Saber Fallah $^{1}$\orcidC{}}

\AuthorNames{Sampo Kuutti, Richard Bowden and Saber Fallah}

\AuthorCitation{Kuutti, S.; Bowden, R.; Fallah, S.}
\address{%
$^{1}$ \quad Connected and Autonomous Vehicles Lab, University of Surrey, Guildford GU2 7XH, UK
; {s.fallah}@surrey.ac.uk\\
$^{2}$ \quad Centre for Vision Speech and Signal Processing, University of Surrey, Guildford GU2 7XH, UK; r.bowden@surrey.ac.uk}% AUTHOR: I confirm this is correct.

\corres{Correspondence: s.j.kuutti@surrey.ac.uk}

\abstract{The use of neural networks and reinforcement learning has become increasingly popular in autonomous vehicle control. However, the opaqueness of the resulting control policies presents a significant barrier to deploying neural network-based control in autonomous vehicles. In this paper, we present a reinforcement learning based approach to autonomous vehicle longitudinal control, where the rule-based safety cages provide enhanced safety for the vehicle as well as weak supervision to the reinforcement learning agent. By guiding the agent to meaningful states and actions, this weak supervision improves the convergence during training and enhances the safety of the final trained policy. This rule-based supervisory controller has the further advantage of being fully interpretable, thereby enabling traditional validation and verification approaches to ensure the safety of the vehicle. We compare models with and without safety cages, as well as models with optimal and constrained model parameters, and show that the weak supervision consistently improves the safety of exploration, speed of convergence, and model performance. Additionally, we show that when the model parameters are constrained or sub-optimal, the safety cages can enable a model to learn a safe driving policy even when the model could not be trained to drive through reinforcement learning alone. }


\begin{document}
	
%%%%%%%%%%%%%%%%%%%%%%%%%%%%%%%%%%%%%%%%%%

\section{Introduction}

Autonomous driving has gained significant attention within the automotive research community in recent years \cite{eskandarian2019research, montanaro2018towards, kuutti2019deep}. The potential benefits in improved fuel efficiency, passenger safety, traffic flow, and ride sharing mean self-driving cars could have a significant impact on issues such as climate change, road safety, and passenger productivity \cite{eskandarian2012handbook, thrun2010toward, DepartmentforTransport2017}. Deep learning techniques have been demonstrated to be powerful tools for autonomous vehicle control, due to their capability to learn complex behaviours from data and generalise these learned rules to completely new scenarios \cite{kuutti2020survey}. These techniques can be divided in two categories based on the modularity of the system. On one hand, modular systems divide the driving task into multiple sub-tasks such as perception, planning, and control and deploy a number of sub-systems and algorithms to solve each of these tasks. On the other hand, end-to-end systems aim to learn the driving task directly from sensory measurements (e.g., camera, radar) by predicting low-level control actions.

End-to-end approaches have recently risen in popularity, due to the ease of implementation and better leverage of the function approximation of Deep Neural Networks (DNNs). However, the high level of opaqueness in DNNs is one of the main limiting factors in the use of neural network-based control techniques in safety-critical applications, such as autonomous vehicles \cite{borg2018safely, burton2017making, varshney2017safety, rajabli2020software}. As the DNNs used to control the autonomous vehicles become deeper and more complex, their learned control policies and any potential safety issues within them become increasingly difficult to evaluate. This is made further challenging by the complex environment in which autonomous vehicles have to operate in, as it is impossible to evaluate the safety of these systems in all possible scenarios they may encounter once deployed \cite{kalra2016driving, wachenfeld2017new, koopman2016challenges}. One class of solutions to introduce safety in machine learning enabled autonomous vehicles is to utilise a modular approach to autonomous driving, where the machine learning systems are used mainly in the decision making layer, whilst low-level control is handled by more interpretable rule-based systems \cite{xu2018zero, wang2019lane, arbabi2020lane}. Alternatively, safety can be guaranteed through redundancy for end-to-end approaches, where machine learning can be used for vehicle motion control with an additional rule-based virtual safety cage acting as a supervisory controller \cite{heckemann2011safe, vom2020fail}. The purpose of the rule-based virtual safety cage is to check the safety of the control actions of the machine learning system, and to intervene in the control of the vehicle if the safety rules imposed by the safety cages are breached. Therefore, during normal operation, the more intelligent machine learning-based controller is in control of the vehicle. However, if the safety of the vehicle is compromised the safety cages can step in and attempt to bring the vehicle back to a safe state through a more conservative rule-based control policy.

This work extends our previously developed Reinforcement Learning (RL) based vehicle following model \cite{kuutti2019end} and virtual safety cages \cite{kuutti2019safe}. We make important extensions to our previous works, by integrating our safety cages into the RL algorithm. The safety cages not only act as a safety function enhancing the vehicle's safety, but are also used to provide weak supervision during training, by limiting the amount of unnecessary exploration and providing penalties to the agent when breached. In this way, the vehicle can be safe during training by avoiding collisions when the RL agent takes unsafe actions. More importantly, the efficiency of the training process is improved, as the agent converges to an optimal control policy with less samples. We also compare our proposed framework on less safe agents with smaller neural networks, and show significant improvement in the final learned policies when used to train these shallow models. Our contributions can be summarised as follows:

\begin{itemize}
	\item We combine the safety cages with reinforcement learning by intervening on unsafe control actions, as well as providing an additional learning signal for the agent to enable safe and efficient exploration.
	\item We compare the effect of the safety cages during training for both models with optimised hyperparameters, as well as less optimised models which may require additional safety considerations.
	\item We test all trained agents without safety cages enabled, in both naturalistic and adversarial driving scenarios, showing that even if the safety cages are only used during training, the models exhibit safer driving behaviour.
	\item We demonstrate that by using the weak supervision from the safety cages during training, the shallow model which otherwise could not learn to drive can be enabled to learn to drive without collisions.
\end{itemize}

The remainder of this paper is structured as follows. Section \ref{sec2} discusses related work and explains the novelty of our approach. Section \ref{sec3} provides the necessary theoretical background for the reader and describes the methodology used for the safety cages and reinforcement learning technique. The results from the simulated experiments are presented and discussed in Section \ref{sec4}. Finally, the concluding remarks are presented in Section \ref{sec5}.
 
%%%%%%%%%%%%%%%%%%%%%%%%%%%%%%%%%%%%%%%%%%
\section{Related Work}\label{sec2}
\subsection{Autonomous Driving}
A brief overview of relevant works in this field is given in this Section. For a more in-depth view of deep learning based autonomous driving techniques, we refer the interested readers to the review in \cite{kuutti2020survey}. One of the earliest works in neural control for autonomous driving was Pomerleau's Autonomous Land Vehicle In a Neural Network (ALVINN) \cite{pomerleau1989alvinn}, which learned to steer an autonomous vehicle by observing images from a front facing camera, using the recorded steering commands of a human driver as training data. Among the first to adapt techniques such as ALVINN to use deep neural networks, was NVIDIA's PilotNet \cite{bojarski2016end}. PilotNet was trained for lane keeping using supervised learning with a total of 72 h of recorded human driving as training data. 

Since then, these works have inspired a number of deep learning techniques, with imitation learning often being the preferred learning technique. For instance, \mbox{Zhang et al. \cite{zhang2016query}} and Pan et al. \cite{pan2018agile} extended the popular Dataset Aggregation (DAgger) \cite{ross2011reduction} imitation learning algorithm to the autonomous driving domain, demonstrating that autonomous vehicle control can be learned from vision. While imitation learning based approaches have shown important progress in autonomous driving \cite{codevilla2018end, bansal2018chauffeurnet, wang2018deep, hecker2018end}, they present limitations when deployed in environments beyond the training distribution \cite{codevilla2019exploring}. These driving models relying on supervised techniques are often evaluated on performance metrics on pre-collected validation datasets \cite{xu2017end}, however low prediction error on offline testing is not necessarily correlated with driving quality \cite{codevilla2018offline}. Even when demonstrating desirable performance during closed-loop testing in naturalistic driving scenarios, imitation learning models often degrade in performance due to distributional shift \cite{ross2011reduction}, unpredictable road users \cite{kuutti2020training}, or causal confusion \cite{de2019causal} when exposed to a variety of driving scenarios.

{However, RL-based techniques have shown promising results for autonomous vehicle applications \cite{kiran2020deep, wu2020battery, wu2020batteryb}. These RL approaches are advantageous for autonomous vehicle motion control, as they can learn general driving rules, which can also adapt to new environments.} Indeed, many recent works have utilised RL for longitudinal control in autonomous vehicles with great success \cite{puccetti2019actor, chae2017autonomous, zhao2017model, li2019ecological, huang2017parameterized}. This is largely due to the fact that longitudinal control can be learned from low-dimensional observations (e.g., relative distances, velocities), which partially overcomes the sample-efficiency problem inherent in RL. Moreover, the reward function for RL is easier to define in the longitudinal control case (e.g., based on safety distances to vehicles in front). For these reasons, we focus on longitudinal control and extend on our previous work on RL-based longitudinal control in a highway driving environment \cite{kuutti2019end}.

\subsection{Safety Cages}

Virtual safety cages have been used in several cyber-physical systems to provide safety guarantees when the controller is not interpretable. The most straightforward application of such safety cages is to limit the possible actions of the controller to ensure the system is bounded to a safe operational envelope. If the controller issues commands that breach the safety cages, the safety cages step in and attempt to recover the system back to a safe state. This type of approach has been used to guarantee the safety of complex controllers in different domains such as robotics \cite{kurien1998model, crestani2015enhancing, haddadin2012making, kuffner2016virtual}, aerospace \cite{polycarpou2004neural}, and automotive \mbox{applications \cite{adler2016safety, jackson2019certified, pek2020using}.} Heckemann et al. \cite{heckemann2011safe} suggested that these safety cages could be used to ensure the safety of black box systems in autonomous vehicles by utilising the vehicle's sensors to monitor the state of the environment, and then limiting the actions of the vehicle in safety-critical scenarios. Demonstrating the effectiveness of this approach, \mbox{Adler et al. \cite{adler2016safety}} proposed five safety cages based on the Automotive Safety Integrity Levels (ASIL) defined by \mbox{ISO26262 \cite{iso26262}} to improve the safety of an autonomous vehicle with machine learning based controllers. Focusing on path planning in urban environments, Yurtsever et al. \cite{yurtsever2020integrating} combined RL with rule-based path planning to provide safety guarantees in autonomous driving. Similar approaches have also been used for highway driving, by combining rule-based systems with machine learning based controllers for enhanced driving safety \cite{likmeta2020combining, baheri2019deep}. 

In our previous work \cite{kuutti2019safe}, we developed two safety cages for highway driving, and demonstrated these safety cages can be used to prevent collisions when the neural network controllers make unpredictable decisions. Furthermore, we demonstrated that the interventions by the safety cages can be used to re-train the neural networks in a supervised learning approach, enabling the system to learn from its own mistakes and further making the controller more robust. However, the main limitation of the safety cage approach was that the re-training happened in an offline manner, where the learning was broken down into three stages: (i) supervised training, (ii) closed-loop evaluation with safety cages, and (iii) re-training using the safety cage interventions as labels for supervised learning.

Here, we extend on this approach by utilising the safety cages to improve the safety of a RL based vehicle motion controller, and using the interventions of the safety cages as weak supervision which enables the system to learn to drive more safely in an online manner. Weak supervision has been shown to improve the efficiency of exploration in \mbox{RL \cite{lee2020weakly}} by guiding the agent towards useful directions during exploration. Here, the weak supervision enhances the exploration process in two ways; the safety cages stop the vehicle from taking unsafe actions thereby eliminating the unsafe part of the action space from the exploration, while also maintaining the vehicle in a safe state and thereby reducing the amount of states that need to be explored. {Reinforcement learning algorithms often struggle to learn efficiently at the beginning of training, since initially the agent is taking largely random actions, and it can take a significant amount of training before the agent starts to take the correct actions which are needed to learn its task. Therefore, by utilising weak supervision to guide the agent to the correct actions and states, the efficiency of the early training stage can be improved.} We show that eliminating the unsafe parts of the exploration space improves convergence during training, which can be a significant advantage considering the low sample efficiency of RL. Furthermore, we show that the safety cages eliminate the collisions that would normally happen during training, which could be a further advantage of our technique, should the training occur in a real-world system where collisions are undesirable.

%%%%%%%%%%%%%%%%%%%%%%%%%%%%%%%%%%%%%%%%%%
\section{Materials and Methods}\label{sec3}
\subsection{Reinforcement Learning}

\end{paracol}

\begin{figure}[H]
\widefigure
%\centering
\includegraphics[width=16 cm]{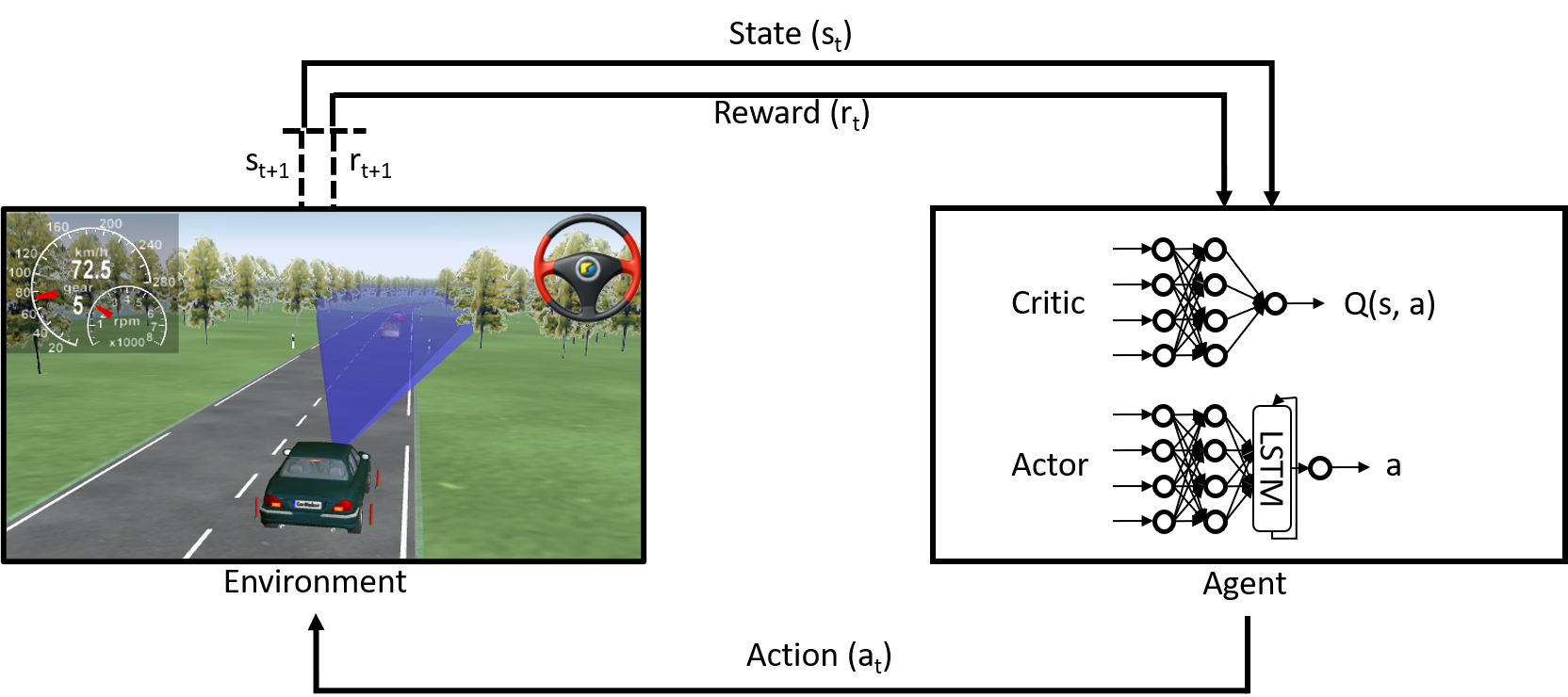}
\caption{Reinforcement learning process.\label{fig:rl}}
\end{figure}
\begin{paracol}{2}
%\linenumbers
\switchcolumn

Reinforcement learning can be formally described as a Markov Decision Process (MDP). The MDP is denoted by a tuple \{$\mathcal{S}, \mathcal{A}, \mathcal{P}, \mathcal{R}$\}, where $\mathcal{S}$ represents the state space, $\mathcal{A}$ represents the the action space, $\mathcal{P}$ denotes the state transition probability model, and $\mathcal{R}$ is the reward function \cite{sutton1998introduction}. As shown in Figure~\ref{fig:rl}, at each time step $t$, the agent takes an action $a_t$ from the possible set of actions $\mathcal{A}$, according to its policy $\pi$ which is a mapping from states $s_t$ to actions $a_t$. Based on the action taken in the current state, the environment then transitions to the next state $s_{t+1}$ according to the transition dynamics $p(s_{t+1}|s_t, a_t)$ as given by the transition probability model $\mathcal{P}$. The agent then observes the new state $s_{t+1}$ and receives a scalar reward $r_t$ according to the reward function $\mathcal{R}$. The aim of the agent in the RL setting is to maximise the total accumulated returns $R_t$:

\begin{equation}
R_t = \sum_{i=t}^{\infty}\gamma^{(i-t)}r(s_i,a_i)
\end{equation}
where $\gamma \in [0,1]$ is the discount factor used to prioritise immediate rewards over \mbox{future rewards.}

\subsection{Deep Deterministic Policy Gradient}

Deep Deterministic Policy Gradient (DDPG) \cite{lillicrap2015continuous} extends the Deterministic Policy Gradient algorithm by Silver et al. \cite{silver2014deterministic} by utilising DNNs for function approximation. It is an actor-critic based off-policy RL algorithm, which can scale to high-dimensional and continuous state and action spaces. The DDPG uses the state-action value function, or Q-function, $Q(s, a)$, which estimates the expected returns after taking an action $a_t$ in \mbox{state $s_t$} under policy $\pi$. Therefore, given a state visitation distribution $\rho^\pi$ under policy $\pi$ in environment $E$ the Q-function is denoted by:

\begin{equation}
Q^\pi(s_t, a_t) = \mathbb{E}_{r_{i \geq t}, s_{i>t}\sim E,a_{i>t} \sim \pi}[R_t|s_t, a_t]
\end{equation}

The Q-function can be estimated by the Bellman equation for deterministic policies as:

\begin{equation}
Q^\pi(s_t, a_t) = \mathbb{E}_{r_t,s_{t+1} \sim E}[r(s_t, a_t) + \gamma Q^\pi (s_{t+1}, \pi(s_{t+1}))]
\end{equation}

As the expectations depend only on the environment, the critic network can be trained off-policy, using transitions from a different stochastic policy with the state visitation distribution $\rho^\beta$. The parameters of the critic network $\theta^Q$ can then be updated by minimising the critic loss $\mathcal{L}_Q$: 

\begin{equation}
\mathcal{L}_Q = \mathbb{E}_{s_t\sim\rho^\beta,r_t\sim E}[(Q(s_t, a_t| \theta^Q) - y_t)^2]
\end{equation}
where
\begin{equation}
y_t = r(s_t, a_t) + \gamma Q(s_{t+1}, \pi(s_{t+1})|\theta^Q)
\end{equation}

The actor network parameters $\theta^\pi$ are then updated using the policy gradient \cite{silver2014deterministic} from the expected returns from a start distribution $J$ with respect to the actor parameters $\theta^\pi$:

\begin{multline}
\triangledown_{\theta^\pi}J \approx \mathbb{E}_{s_t\sim\rho^\beta}[\triangledown_{\theta^\pi}Q(s, a|\theta^Q)|_{s=s_t,a=\pi(s_t|\theta^\pi)}] \\
= \mathbb{E}_{s_t\sim\rho^\beta}[\triangledown_{a}Q(s,a|\theta^Q)|_{s=s_t, a=\pi(s_t)} \triangledown_{\theta^\pi}\pi(s|\theta^\pi)|_{s=s_t}]
\end{multline}

For updating the networks, mini-batches are drawn from a replay memory $\mathcal{D}$, which is a finite sized buffer storing state transitions $e = [s_t, a_t, r_t, s_{t+1}]$. To avoid divergence and improve stability of training, DDPG utilises target networks \cite{mnih2013playing}, which copy the parameters of the actor and critic networks. These target networks, target actor $\pi ' (s|\theta^{\pi '})$ and target critic network $Q'(s,a|\theta^{Q'})$, are updated slowly based on the learned network parameters to improve stability:

\begin{equation}
\theta' \leftarrow \tau \theta + (1 - \tau)\theta'
\end{equation}
where $\tau \ll 1$ is the mixing factor, a hyperparameter controlling the speed of target network updates.

To encourage the agent to explore the possible actions for continuous action spaces, noise is added to the actions of the deterministic policy $\pi(s_t|\theta^\pi)$. This exploration policy $\pi^e(s_t)$, samples noise from a noise process $\mathcal{N}$ which is added to the actor policy:

\begin{equation}
\pi^e(s_t) = \pi(s_t|\theta_t^\pi) + \mathcal{N}
\end{equation}
Here, the chosen noise process $\mathcal{N}$ is the Ornstein-Uhlenbeck process \cite{uhlenbeck1930theory}, which generates temporally correlated noise for efficient exploration in physical control problems.

\subsection{Safety Cages}

Virtual safety cages provide interpretable rule-based safety for complex cyber-physical systems. The purpose of these safety cages is to limit the actions of the system to a safe operational envelope. The simple way to achieve this, would be to limit the upper or lower limits of the system's action space. However, by using run-time monitoring to observe the state of the environment, the safety cages can dynamically select the control limits based on the current states. Therefore, the system can be limited in its possible courses of action when faced with a safety-critical scenario, such as a near-accident situation on a highway. We utilise our previously presented safety cages \cite{kuutti2019safe}, which limit the longitudinal control actions of a vehicle based on the Time Headway (TH) and Time-To-Collision (TTC) relative to the vehicle in front. The TH and TTC metrics represent the risk of potential forward collision with the vehicle in front, and are calculated as:

\begin{equation}
TH = \frac{x_{rel}}{v}
\end{equation}
\begin{equation}
TTC = \frac{x_{rel}}{v_{rel}}
\end{equation}
where $x_{rel}$ is the distance between the two vehicles in m, $v$ is the velocity of the host vehicle in m/s, and $v_{rel}$ is the relative velocity between the two vehicles in m/s.

The TTC and TH metrics were chosen as the states monitored by the safety cages as they represent the risk of potential collision with the vehicle in front, thereby providing effective safety measurements for our vehicle following use-case. We utilise two metrics as the TTC and TH provide complimentary information; the TTC measures time to a forward collision assuming both vehicles continue at their current speeds, whilst TH measures distance to the vehicle in front in time and makes no assumptions about the lead vehicle's actions. For example, when the host vehicle is driving significantly faster than the vehicle in front, as the distance between the vehicles gets closer the TTC approaches zero and correctly captures the risk of a forward collision. However, in a scenario where both vehicles are driving close to each other but at the same speed, the TTC will not signal a high risk of collision even though in this scenario if the lead vehicle begins to break, the two vehicles would be in a likely collision. In such a scenario, the two vehicles will have a low headway, therefore monitoring the TH will correctly inform the safety monitors of a collision risk.

The risk levels for both safety cages are as defined in \cite{kuutti2019safe}, where the aim was to identify potential collisions in time to prevent them, whilst minimising unnecessary interventions on the control of the vehicle. The different risk levels and associated minimum braking values are illustrated in Figure~\ref{fig:cages}. For each safety cage, there are three risk levels for which the safety cages will enforce a minimum braking value on the vehicle, with higher risk levels using increased rate of braking relative to the associated safety metric. When the vehicle is in the low risk region, no minimum braking is necessary and the RL agent is in full control of the vehicle. The minimum braking values enforced by the safety cages can be formally defined as shown in (\ref{eqn:sc_th})--(\ref{eqn:sc_ttc}). The braking value is normalised to the range [0, 1] where 0 is no braking and 1 is maximum braking value. In this framework, both safety cages provide a recommended braking value, which is then compared to the current braking action from the RL agent. The final braking value used for the vehicle motion control, $b$, is then chosen as the largest braking value between the two safety cages and the RL agent as given by (\ref{eqn:min_b}).

\begin{equation}\label{eqn:sc_th}
b_{TH} = 
\begin{cases}
0.0, & \text{for TH $>$ 1.6} \\
-0.5TH + 1.0, & \text{for 1.0 $<$ TH $\leq$ 1.6} \\
-1.0TH + 1.5, & \text{for 0.5 $<$ TH $\leq$ 1.0} \\
1.0, & \text{for TH $\leq$ 0.5} \\
\end{cases}
\end{equation}
\begin{equation}\label{eqn:sc_ttc}
b_{TTC} = 
\begin{cases}
0.0, & \text{for TTC $>$ 2.5} \\
-0.5TTC + 1.25, & \text{for 1.5 $<$ TTC $\leq$ 2.5} \\
-1.0TTC + 2.0, & \text{for 1.0 $<$ TTC $\leq$ 1.5} \\
1.0, & \text{for TTC $\leq$ 1.0} \\
\end{cases}
\end{equation}
\begin{equation}\label{eqn:min_b}
b = max(b_{TH}, b_{TTC}, b_{RL})
\end{equation}
where $b_{TH}$ is the minimum braking value from the TH safety cage, $b_{TTC}$ is the minimum braking value from the TTC safety cage, and $b_{RL}$ is the current braking value from the \mbox{RL agent.}

% start a new page without indent 4.6cm
%\clearpage
\end{paracol}
\nointerlineskip
\begin{figure}[H]
	\widefigure
	%\centering
	\includegraphics[width=18 cm]{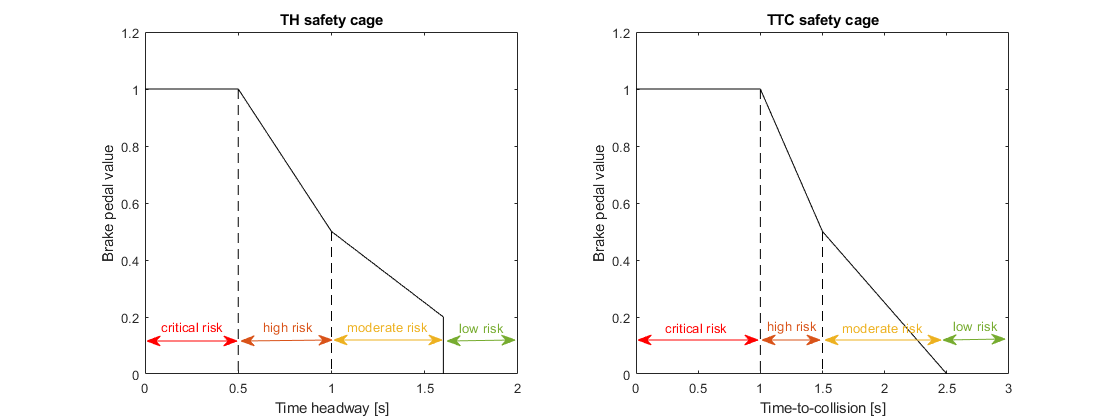}
	\caption{Risk-levels and minimum braking values for each safety cage.}
	\label{fig:cages}
\end{figure}
\begin{paracol}{2}
%\linenumbers
\switchcolumn

\subsection{Highway Vehicle Following Use-Case}

The vehicle following use-case was framed as a scenario on a straight highway, with two vehicles travelling in a single lane. The host vehicle controlled by RL is the follower vehicle, and its aim is to maintain a 2 s headway from the lead vehicle. The lead vehicle velocities were limited to $v_{lead} \in$ [17, 40] m/s, and coefficient of friction values were chosen from the range [0.4, 1.0] for each episode. During training, the lead vehicle's acceleration was limited to $\dot{v}_{lead} \in$ [$-$2, 2] m/s\textsuperscript{2}, except for emergency braking manoeuvrers, which occurred on average once an hour, which used an acceleration in the range \mbox{$\dot{v}_{lead} \in$ [$-$6, $-$3] m/s\textsuperscript{2}}. The output from the RL agent is the gas and brake pedal values, which are continuous action values used to control the vehicle. As in \cite{kuutti2019end}, a neural network is used to estimate the longitudinal vehicle dynamics, by inferring the vehicle response to the pedal actions from the RL agent. This neural network acts as a type of World \mbox{Model \cite{ha2018world}}, providing an estimation of the simulator environment. This has the advantage that the neural network can be deployed on the same GPU as the RL network during training, thereby speeding up training time significantly. The World Model was trained with 2,364,041 time-steps from the IPG CarMaker simulator under different driving policies combining a total of 45 h of simulated driving. This approach was shown in \cite{kuutti2019end} to speed up training by up to a factor of 20, compared to training with the IPG CarMaker simulator. However, to ensure the accuracy of all results, we also evaluate all trained policies in IPG CarMaker (Section~\ref{sec4.1}).

\subsection{Training}

The DDPG model is trained in the vehicle following environment for 5000 episodes, where each episode lasts up to 5 min or until a collision occurs. The simulation is sampled at 25 Hz, therefore each time-step has a duration of 40 ms. The training parameters of the DDPG were tuned heuristically, and the final values can be found in Table \ref{tbl:netarch}. The critic uses a single hidden layer, followed by the output layer estimating the $Q$ value. The actor network utilises 3 hidden feedforward layers, followed by a Long Short-Term Memory (LSTM) \cite{hochreiter1997long} and then the action layer. {The actor network outputs the vehicle control action, for which the action space is represented by a single continuous value $a_t \in$ [$-$1, 1], where positive values represent the use of the gas pedal and negative values represent the use of the brake pedal. The observations of the agent are composed of 4 continuous state-values, which are the host vehicle velocity $v$, host vehicle \mbox{acceleration $\dot{v}$}, relative velocity $v_{rel}$, and time headway $TH$, such that $s_t = [v, \dot{v}, v_{rel}, TH]^T$.} To enable the LSTM to learn temporal correlations, the mini-batches for training were sampled as consecutive time-steps, with the LSTM cell state reset between each training update. To encourage the agent to learn a safe vehicle following policy, a reward function based on its current headway and headway derivative was defined in \cite{kuutti2019end} based on the reward function by Desjardins \& \mbox{Chaib-Draa \cite{desjardins2011cooperative}}, as shown in Figure~\ref{fig:reward}. The agent gains the maximum reward when it is close to the target headway of 2 s, whilst straying further from the target headway results in smaller rewards. The headway derivative is used in the reward function to encourage the vehicle to move towards the target headway, by giving small positive rewards as it moves closer to the target and penalising the agent when it is moving further away from the target region. For further comparison, we compare training the model with an additional penalty for breaching the safety cages, such that the final reward is given \mbox{as follows:}

\begin{equation}
r_t = r_{th} + r_{sc}
\end{equation}
where $r_t$ is the reward for time-step $t$, $r_{th}$ is the headway based reward function as shown in Figure~\ref{fig:reward}, and $r_{sc}$ is the safety cages penalty equal to -0.1 if the safety cage is breached and 0 otherwise.

\begin{specialtable}[H]
	\caption{Network hyperparameters.}
	\label{tbl:netarch}
\setlength{\cellWidtha}{\columnwidth/2-2\tabcolsep+0.8in}
\setlength{\cellWidthc}{\columnwidth/2-2\tabcolsep-0.8in}
\scalebox{1}[1]{\begin{tabularx}{\columnwidth}{
>{\PreserveBackslash\centering}m{\cellWidtha}
>{\PreserveBackslash\centering}m{\cellWidthc}}

\toprule

			\textbf{Parameter} & \textbf{Value} \\
			
	\midrule

			Mini-batch size & 64 \\
			Hidden neurons in feedforward layers & 50 \\
			LSTM units & 16 \\
			Discount factor $\gamma$ & 0.99 \\
			Actor learning rate $\eta_{\pi}$ & 10\textsuperscript{$-$4} \\
			Critic learning rate $\eta_{q}$ & 10\textsuperscript{$-$2} \\
			Replay memory size & 10\textsuperscript{6} \\
			Mixing factor $\tau$ & 10\textsuperscript{$-$3} \\
			Initial exploration noise scale & 1.0 \\
			Max gradient norm for clipping & 0.5 \\
			Exploration noise decay & 0.997 \\
			Exploration mean $\mu$ & 0.0 \\
			Exploration scale factor $\theta$ & 0.15 \\
			Exploration variance $\sigma$ & 0.2 \\
		\bottomrule

\end{tabularx}} 
\end{specialtable}

% Reward func fig. here
\begin{figure}[H]
	\includegraphics[width=0.5\textwidth]{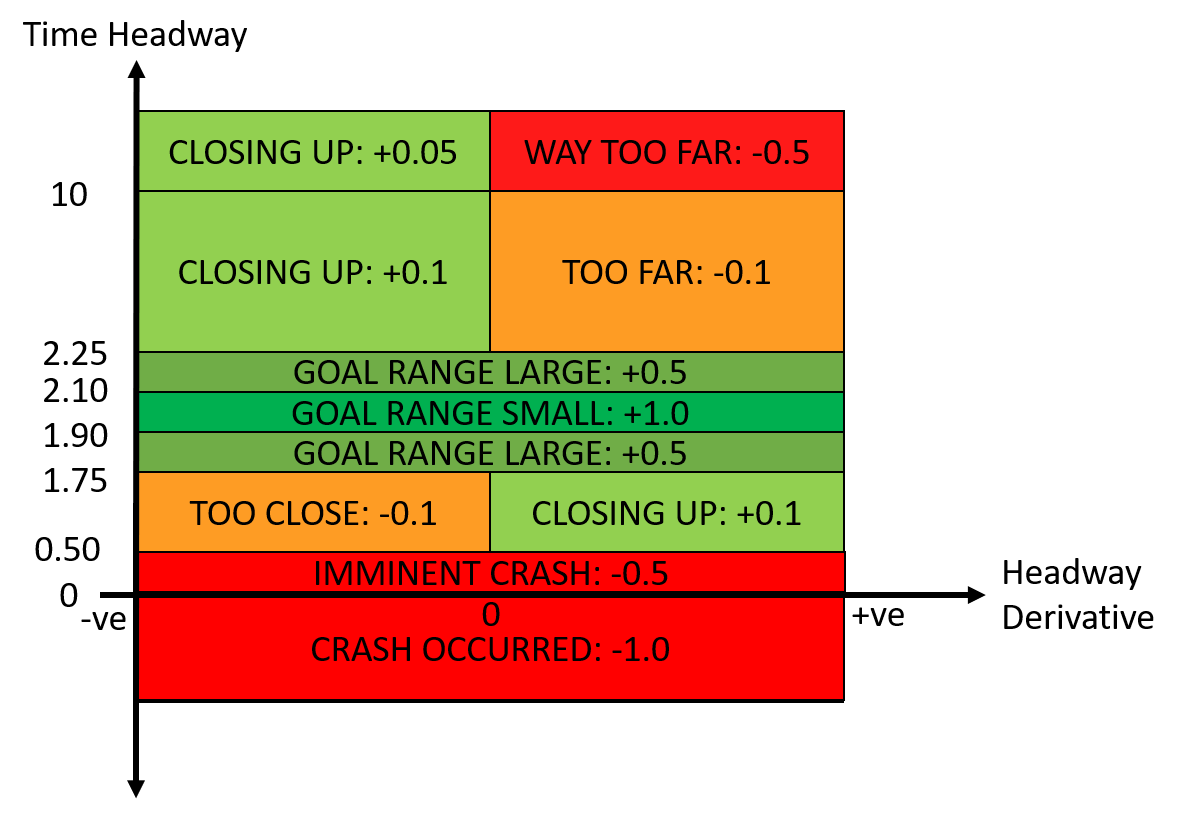}
	\caption{Time headway based reward function for vehicle following.}
	\label{fig:reward}
\end{figure}

The episode rewards during training can be seen in Figure~\ref{fig:training}, where three models are compared. The three models are DDPG only, DDPG+SC which is DDPG with safety cages, and DDPG+SC (no penalty) which is the DDPG with safety cages but without the $r_{sc}$ penalty. As can be seen, the DDPG+SC model has lower rewards at the beginning of training as it receives additional penalties compared to the other two models. However, after the initial exploration the DDPG+SC is the first model to reach the optimal rewards per episode ($\sim$7500 rewards), demonstrating improved convergence. Comparing the DDPG+SC models with and without penalties from the safety cages shows the model with the penalties converges to the optimal solution sooner, suggesting the penalty improves convergence during training. An additional benefit of the safety cages here is the safety of exploration, as the DDPG model collided 30 times during training, whilst the DDPG+SC model had no collisions during training. However, it can be seen that all three models converge to the same level of performance, therefore no significant difference in the trained policies can be concluded from the training rewards alone. 

% Training Fig. here
\begin{figure}[H]
	\includegraphics[width=0.5\textwidth]{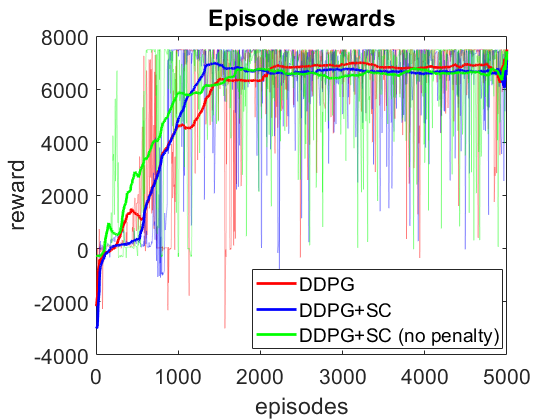}
	\caption{Episode rewards during DDPG training. Darker lines represent the moving average reward plot, whilst the actual reward values are seen in the transparent region.}
	\label{fig:training}
\end{figure}

As an additional investigation of the effect of the safety cages on less safe control policies, we train two further models utilising smaller neural networks with constrained parameters. These models use the same parameters as in Table \ref{tbl:netarch}, except they only have 1 single hidden layer with 50 neurons and no LSTM layer. We refer to these models as Shallow DDPG and Shallow DDPG+SC. It should be noted that the parameters of these models were not tuned for better performance, and indeed sub-optimal parameters were chosen on purpose to enable better insight into the effect of the safety cages in unsafe systems. The episode rewards for the two shallow models during training are shown \mbox{in Figure~\ref{fig:training2}}. As can be seen, these two models have a more significant difference in training performance. The Shallow DDPG struggles to learn a feasible training policy, whilst the Shallow DDPG+SC learns to drive without collisions, although at a lower level of overall performance compared to the deeper models.

% Training Fig. here
\begin{figure}[H]
	\includegraphics[width=0.5\textwidth]{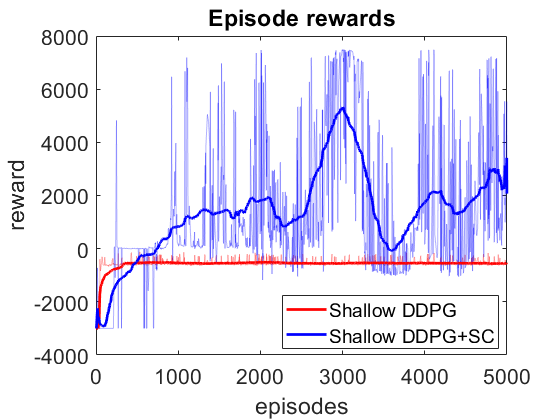}
	\caption{Episode rewards during DDPG training of shallow models. Darker lines represent the moving average reward plot, whilst the actual reward values are seen in the transparent region.}
	\label{fig:training2}
\end{figure}

%%%%%%%%%%%%%%%%%%%%%%%%%%%%%%%%%%%%%%%%%%
\section{Results}\label{sec4}

To investigate the performance of the learned control policies, we evaluate the vehicle follower models in various highway driving scenarios. We utilise two types of testing for this evaluation. Naturalistic testing tests the control policies in typical driving scenarios, giving an idea of how the control policies perform in everyday driving. Adversarial testing utilises an adversarial agent to create safety-critical scenarios, showing how the vehicle performs in dangerous edge cases where collisions are likely to occur. The controller performance in both types of scenario is important, since most driving scenarios on the road fall into naturalistic driving the controller must be able to drive efficiently and safely in these scenarios, however the controller must also be able remain safe in dangerous edge cases in order to avoid collisions. To enable better analysis of the performance of the RL-based control policies, no safety cages are used during testing so the vehicle follower models must depend on their own learned knowledge to keep the vehicle safe. This also enables better understanding on the effect of using the safety cages during training on the final learned control policy.

\subsection{Naturalistic Testing}\label{sec4.1}

For the naturalistic driving, similar lead vehicle behaviours were used to those during training, with velocities in the range [17, 40] m/s and acceleration [$-$2, 2] m/s\textsuperscript{2}. The exception to this was the harsh braking manoeuvres which occurred, on average, once an hour with deceleration [$-$6, $-$3] m/s\textsuperscript{2}. At the start of the episode, the coefficient of friction is randomly chosen in the range $[0.4, 1.0]$ and each episode lasts until 5 min has passed or a collision occurs. For each driving model, a total of 120 test scenarios were completed, totalling up to 10 h of testing. All driving for these tests occurred in the IPG CarMaker simulation environment to ensure accuracy of the results. Two types of baselines are provided for comparison; the IPG Driver is the default driver in the CarMaker Simulator and A2C is the Advantage Actor Critic \cite{mnih2016asynchronous} based vehicle follower model in \cite{kuutti2019end}.

The results from the naturalistic driving scenarios are summarised in Table \ref{tbl:simresults}. The table shows the RL based models outperform the default IPG Driver, with the exception of the shallow models. The results demonstrate that both DDPG based models outperform the previous A2C-based vehicle follower model. However, comparing the DDPG and DDPG+SC models shows the benefit of using the safety cages during RL training. While in most scenarios the two models have similar performance (the mean values seen are approximately equal), the minimum headway by the DDPG+SC during testing is higher, showing it can maintain a safer distance from the lead vehicle. However, as both models can maintain a safe distance without collisions this difference is not significant by itself. Therefore, investigating the difference between the Shallow DDPG and Shallow DDPG+SC models provides further insight into the role the safety cages play in supervision during RL training. Similar to the training rewards, the shallow models show a more extreme difference between the two models. The results show the Shallow DDPG model without safety cages fails to learn to drive safely, whilst the Shallow DDPG+SC model avoids collisions safely, although it comes relatively close to collisions with a minimum time headway at 0.79 s. This shows the benefit of the safety cages in guiding the model towards a safe control policy during training.

% start a new page without indent 4.6cm
%\clearpage
\end{paracol}
\nointerlineskip
\begin{specialtable}[H]
\widetable
	\caption{10-hour driving test under naturalistic driving conditions in IPG CarMaker.}
	\label{tbl:simresults}
\setlength{\cellWidtha}{\columnwidth/7-2\tabcolsep+0.0in}
\setlength{\cellWidthb}{\columnwidth/7-2\tabcolsep+0.0in}
\setlength{\cellWidthc}{\columnwidth/7-2\tabcolsep+0.0in}
\setlength{\cellWidtha}{\columnwidth/7-2\tabcolsep+0.0in}
\setlength{\cellWidthb}{\columnwidth/7-2\tabcolsep+0.0in}
\setlength{\cellWidthc}{\columnwidth/7-2\tabcolsep+0.0in}
\setlength{\cellWidtha}{\columnwidth/7-2\tabcolsep+0.0in}
\scalebox{1}[1]{\begin{tabularx}{\columnwidth}{
>{\PreserveBackslash\centering}m{\cellWidtha}
>{\PreserveBackslash\centering}m{\cellWidthb}
>{\PreserveBackslash\centering}m{\cellWidtha}
>{\PreserveBackslash\centering}m{\cellWidthb}
>{\PreserveBackslash\centering}m{\cellWidtha}
>{\PreserveBackslash\centering}m{\cellWidthb}
>{\PreserveBackslash\centering}m{\cellWidthc}}

\toprule

		\textbf{Parameter} & \textbf{IPG Driver} & \textbf{A2C} \cite{kuutti2019end} & \textbf{DDPG} & \textbf{DDPG+SC} & \textbf{Shallow DDPG} & \textbf{Shallow DDPG+SC} \\
\midrule

		min. x\textsubscript{rel} [m] & 10.737 & 7.780 & 15.252 & 13.403 & 0.000 & 5.840 \\
		mean x\textsubscript{rel} [m] & 75.16 & 58.01 & 58.19 & 58.24 & 41.45 & 59.34 \\
		max. v\textsubscript{rel} [m/s] & 13.90 & 7.89 & 10.74 & 9.33 & 13.43 & 6.97 \\
		mean v\textsubscript{rel} [m/s] & 0.187 & 0.0289 & 0.0281 & 0.0271 & 4.59 & 0.0328 \\
		min. TH [s] & 1.046 & 1.114 & 1.530 & 1.693 & 0.000 & 0.787 \\
		mean TH [s] & 2.546 & 2.007 & 2.015 & 2.015 & 1.313 & 2.034 \\
		collisions & 0 & 0 & 0 & 0 & 120 & 0 \\ 
	\bottomrule

\end{tabularx}} 
\end{specialtable}
\begin{paracol}{2}
%\linenumbers
\switchcolumn

\subsection{Adversarial Testing}

Utilising machine learning to expose weaknesses in safety-critical cyber-physical systems has been shown to be an effective method for finding failure cases \mbox{effectively \cite{corso2020survey, riedmaier2020survey}}. We utilise the Adversarial Testing Framework (ATF) presented in \cite{kuutti2020training}, which utilised an adversarial agent trained through RL to expose over 11,000 collision cases in machine learning based autonomous vehicle control systems. The adversarial agent is trained through A2C \cite{mnih2016asynchronous} with a reward function $r_A$ based on the inverse headway:

\begin{equation}
r_A = min\left(\frac{1}{TH}, 100\right)
\end{equation}
This reward function encourages the adversarial agent to minimise the headway and make collisions happen, while capping the reward at 100 ensures that the reward does not tend to infinity as the headway reaches zero.

As this lead vehicle used in the adversarial testing can behave very differently to those seen during training, this testing focuses on investigating the models' generalisation capability as well as their response to hazardous scenarios. Each DDPG model is tested under two different velocity ranges; the first limits the lead vehicle's velocity to the same as the training scenarios with $v_{lead} \in [17, 40]$ m/s, and the second uses a lower velocity range which enables the ATF to expose collisions more easily at a velocity range of \mbox{$v_{lead} \in [12, 30]$ m/s.} For each model, 3 different adversarial agents were trained, such that results can be averaged between these 3 training runs. The minimum episode TH during training can be seen for both deep models over the 2500 training episodes in Figures~\ref{fig:at_deep} and \ref{fig:at_deep_low}. These tests show that both deep models can maintain a safe distance from the lead vehicle even when the lead vehicle is attempting to cause collisions intentionally. Although a slight difference in the two models can be seen, as the DDPG+SC model has a slightly higher headway on average as well as significantly less variance. However as both deep models remain at a safe distance from the adversarial agent, these models can be considered safe even in safety-critical edge cases. Comparing the two shallow models in Figures~\ref{fig:at_shallow} and \ref{fig:at_shallow_lw}, a more significant difference can be seen. While both models are worse in performance than the deep models, the Shallow DDPG is significantly easier to exploit than the Shallow DDPG+SC model. The Shallow DDPG model continues to cause collisions during the adversarial testing, whilst the Shallow DDPG+SC model remains at a safer distance. In the training conditions, the Shallow DDPG+SC remains relatively safe, with no decrease in the minimum headway during the training of the adversarial agent, although it can be seen that the variance increases as the training progresses. In the lower velocity case, the Shallow DDPG+SC still avoids collisions, but the adversarial agent is able to reduce the minimum headway significantly better. This shows that the safety cages have helped the model learn a significantly more robust control policy, even when the model uses sub-optimal parameters. Without the additional weak supervision from the safety cages, it can be seen that these shallow models would not have been able to learn a reasonable driving policy. Therefore, the weak-supervision by the safety cages can be used to train models with sub-optimal parameters. In addition, for models with optimal parameters they provide improved convergence during training and slightly improved safety in the final trained policy.

\begin{figure}[H]
	\includegraphics[width=0.45\textwidth]{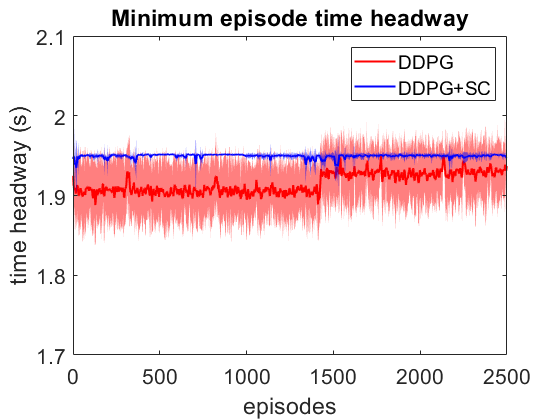}
	\caption{Comparison of the deep vehicle following agents’ minimum TH per episode over adversarial training runs with lead vehicle velocity limits $v_{lead} \in [17, 40]$ m/s. Averaged over 3 runs,
		with standard deviation shown in shaded colour.}
	\label{fig:at_deep}
\end{figure}

\begin{figure}[H]
	\includegraphics[width=0.45\textwidth]{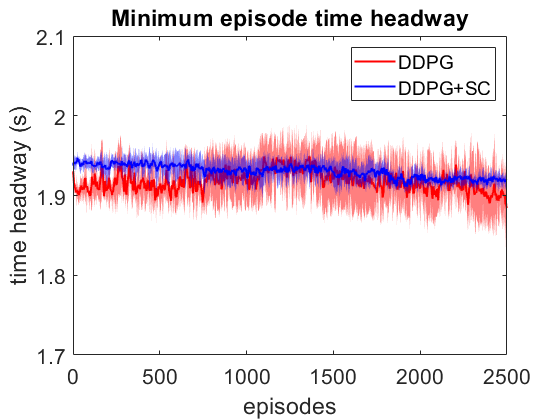}
	\caption{Comparison of the deep vehicle following agents’ minimum TH per episode over adversarial training runs with lead vehicle velocity limits $v_{lead} \in [12, 30]$ m/s. Averaged over 3 runs,
		with standard deviation shown in shaded colour.}
	\label{fig:at_deep_low}
\end{figure}

\begin{figure}[H]
	\includegraphics[width=0.45\textwidth]{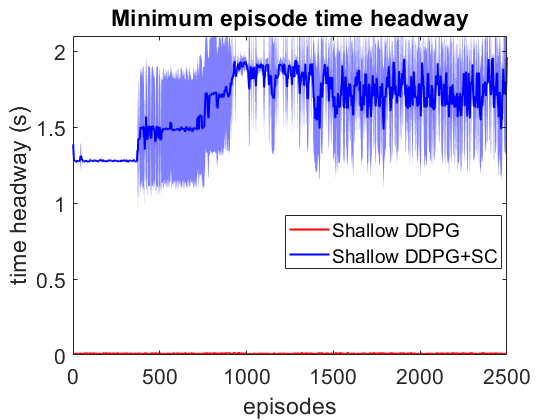}
	\caption{Comparison of the shallow vehicle following agents’ minimum TH per episode over adversarial training runs with lead vehicle velocity limits $v_{lead} \in [17, 40]$ m/s. Averaged over 3 runs,
		with standard deviation shown in shaded colour.}
	\label{fig:at_shallow}
\end{figure}

\begin{figure}[H]
	\includegraphics[width=0.45\textwidth]{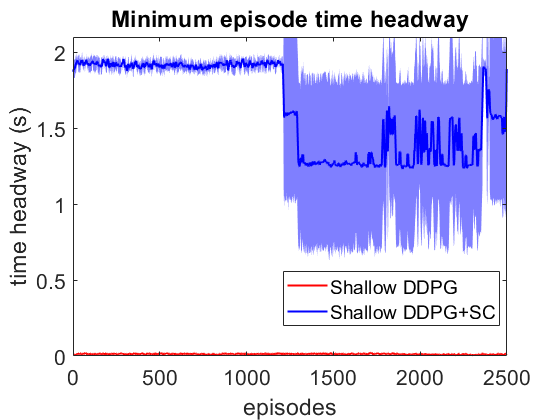}
	\caption{Comparison of the shallow vehicle following agents’ minimum TH per episode over adversarial training runs with lead vehicle velocity limits $v_{lead} \in [12, 30]$ m/s. Averaged over 3 runs,
		with standard deviation shown in shaded colour.}
	\label{fig:at_shallow_lw}
\end{figure}

%%%%%%%%%%%%%%%%%%%%%%%%%%%%%%%%%%%%%%%%%%
\section{Conclusions}\label{sec5}

In this paper, a reinforcement learning technique combining rule-based safety cages was presented. The safety cages provide a safety mechanism for the autonomous vehicle in case the neural network-based controller makes unsafe decisions, thereby enhancing the safety of the vehicle and providing interpretability in the vehicle motion control system. In addition, the safety cages are used as weak supervision during training, by guiding the agent towards useful actions and avoiding dangerous states. 

We compared the model with safety cages to a model without them, and show improvements in safety of exploration, speed of convergence, and the safety of the final control policy. {In addition to improved training efficiency, simulated testing scenarios demonstrated} that even with the safety cages disabled, the model which used them during training has learned a safer control policy {by maintaining a minimum headway of 1.69 s in a safety-critical scenario, compared to 1.53 s without safety cage training}. We additionally tested the proposed approach on shallow models with constrained parameters, and {showed that the shallow model with safety cage training was able to drive without collisions, whilst the shallow model without safety cage training collided in every test scenario. These results }demonstrate that the safety cages enabled the shallow models to learn a safe control policy while otherwise the shallow models were not able to learn a feasible driving policy. This showed that the safety cages add beneficial supervision during training, enabling the model to learn from the environment more effectively. 

Therefore, this work provides an effective way to combine reinforcement learning based control with rule-based safety mechanisms not only to improve the safety of the vehicle, but also incorporating weak supervision in the training process for improved convergence and performance.

{This work opens up multiple potential avenues for future work. The use-case in this study was a simplified vehicle following scenario. However, extending the safety cages to consider both longitudinal and lateral control actions, as well as potential objects on other lanes, would allow the technique to be applied to more complex use-cases such as urban driving. Moreover, comparing the use of the weak supervision for different use-cases or learning algorithms (e.g., on-policy vs. off-policy RL) would help with understanding the most efficient use of weak supervision in reinforcement learning. Furthermore, extending the reinforcement learning agent to use more high dimensional inputs, such as images, would allow investigation into how the increased speed of convergence helps in cases where the sample inefficient reinforcement learning algorithms struggle. Finally, using the safety cages presented here in real-world training could better demonstrate the benefit in both safety and efficiency of exploration, compared to the simulated scenario presented in this work.}
%%%%%%%%%%%%%%%%%%%%%%%%%%%%%%%%%%%%%%%%%%
\vspace{6pt} 

%%%%%%%%%%%%%%%%%%%%%%%%%%%%%%%%%%%%%%%%%%
%% optional
%\supplementary{The following are available online at \linksupplementary{s1}, Figure S1: title, Table S1: title, Video S1: title.}

% Only for the journal Methods and Protocols:
% If you wish to submit a video article, please do so with any other supplementary material.
% \supplementary{The following are available at \linksupplementary{s1}, Figure S1: title, Table S1: title, Video S1: title. A supporting video article is available at doi: link.} 

%%%%%%%%%%%%%%%%%%%%%%%%%%%%%%%%%%%%%%%%%%
\authorcontributions{Conceptualization, S.K., R.B. and S.F.; methodology, S.K.; software, S.K.; validation, S.K.; formal analysis, S.K.; investigation, S.K.; resources, S.K.; data curation, S.K.; writing---original draft preparation, S.K.; writing---review and editing, R.B and S.F.; visualization, S.K.; supervision, R.B. and S.F.; project administration, R.B. and S.F.; funding acquisition, R.B. and S.F. All authors have read and agreed to the published version of the manuscript.}

\funding{This research was supported by the U.K.-Engineering and Physical Sciences Research Council (EPSRC) under Grant EP/R512217.}

\institutionalreview{Not applicable.}

\informedconsent{Not applicable.}

\dataavailability{Data sharing not applicable.}

\conflictsofinterest{The authors declare no conflict of interest.}

%%%%%%%%%%%%%%%%%%%%%%%%%%%%%%%%%%%%%%%%%%
%% Only for journal Encyclopedia
%\entrylink{The Link to this entry published on the encyclopedia platform.}
\newpage
%%%%%%%%%%%%%%%%%%%%%%%%%%%%%%%%%%%%%%%%%%
\end{paracol}
\reftitle{References}

\end{document}